\documentclass[twoside,11pt]{article}

\usepackage{times}
\usepackage{listings}
\usepackage{lscape}
\usepackage[abbrvbib,preprint]{jmlr2e}

\usepackage{amsmath}
\usepackage{amssymb}
\usepackage{mathtools}
\usepackage{dsfont}
\usepackage{lipsum}
\usepackage{bm,bbm}
\usepackage{bbding}
\usepackage{paralist}
\usepackage{xspace}
\usepackage[utf8]{inputenc} 
\usepackage[T1]{fontenc}    
\usepackage{hyperref}       
\usepackage{url}            
\usepackage{amsfonts}       
\usepackage{nicefrac}       
\usepackage{microtype} 
\usepackage{multirow}
\usepackage{graphicx}
\usepackage{ragged2e}
\usepackage[most]{tcolorbox}
\definecolor{darkerlogocolor}{RGB}{20, 0, 145}
\newtcolorbox{ttcolorbox}[1][]{colframe=darkerlogocolor, colback=darkerlogocolor!4!white, title=#1}
\usepackage[font=small,labelfont=bf]{caption}
\usepackage{subfloat}
\usepackage{float}
\usepackage{algorithm,algorithmic}
\usepackage{wrapfig}
\usepackage{booktabs}
\usepackage{verbatim}
\usepackage{setspace}
\usepackage{tikz}
\usepackage{tcolorbox}
\definecolor{mydarkblue}{rgb}{0,0.08,0.45}
\hypersetup{
    colorlinks,
    breaklinks,
    linkcolor = mydarkblue,
    citecolor = mydarkblue,
    urlcolor  = black,
} 

\def \x {\mathbf{x}}
\def \y {\mathbf{y}}

\def \z {\mathbf{z}}

\def \R {\mathbb{R}}

\def \Y {\mathcal{Y}}

\def \X {\mathcal{X}}

\renewcommand{\algorithmicrequire}{ \textbf{Input:}}      
\renewcommand{\algorithmicensure}{ \textbf{Output:}}     

\DeclareMathOperator*{\argmax}{arg\,max}

\usepackage{mathtools}
\usepackage{appendix}
 
\let\norm\undefined
\DeclarePairedDelimiter\norm{\lVert}{\rVert}

\newtheorem{myThm}{Theorem}

\newtheorem{myLemma}{Lemma}

\newtheorem{myProp}{Proposition}

\newtheorem{myAssum}{Assumption}

\newtheorem{myDef}{Definition}

\newtheorem{myRemark}{Remark}

\def \R {\mathbb{R}}

\def \X {\mathcal{X}}
\def \Y {\mathcal{Y}}
\def \x {\mathbf{x}}
\def \y {\mathbf{y}}

\def \epsilon {\varepsilon}

\definecolor{DSgray}{cmyk}{0,1,0,0}

\usepackage[textsize=tiny]{todonotes}
\usepackage{prettyref}

\newcommand{\savehyperref}[2]{\texorpdfstring{\hyperref[#1]{#2}}{#2}}
\newrefformat{eq}{\savehyperref{#1}{Eq.~(\ref*{#1})}}
\newrefformat{eqn}{\savehyperref{#1}{Equation~\ref*{#1}}}
\newrefformat{lem}{\savehyperref{#1}{Lemma~\ref*{#1}}}
\newrefformat{lemma}{\savehyperref{#1}{Lemma~\ref*{#1}}}
\newrefformat{def}{\savehyperref{#1}{Definition~\ref*{#1}}}
\newrefformat{line}{\savehyperref{#1}{Line~\ref*{#1}}}
\newrefformat{thm}{\savehyperref{#1}{Theorem~\ref*{#1}}}
\newrefformat{corr}{\savehyperref{#1}{Corollary~\ref*{#1}}}
\newrefformat{cor}{\savehyperref{#1}{Corollary~\ref*{#1}}}
\newrefformat{sec}{\savehyperref{#1}{Section~\ref*{#1}}}
\newrefformat{app}{\savehyperref{#1}{Appendix~\ref*{#1}}}
\newrefformat{appendix}{\savehyperref{#1}{Appendix~\ref*{#1}}}
\newrefformat{assum}{\savehyperref{#1}{Assumption~\ref*{#1}}}
\newrefformat{ex}{\savehyperref{#1}{Example~\ref*{#1}}}
\newrefformat{fig}{\savehyperref{#1}{Figure~\ref*{#1}}}
\newrefformat{alg}{\savehyperref{#1}{Algorithm~\ref*{#1}}}
\newrefformat{rem}{\savehyperref{#1}{Remark~\ref*{#1}}}
\newrefformat{conj}{\savehyperref{#1}{Conjecture~\ref*{#1}}}
\newrefformat{prop}{\savehyperref{#1}{Proposition~\ref*{#1}}}
\newrefformat{proto}{\savehyperref{#1}{Protocol~\ref*{#1}}}
\newrefformat{prob}{\savehyperref{#1}{Problem~\ref*{#1}}}
\newrefformat{claim}{\savehyperref{#1}{Claim~\ref*{#1}}}
\newrefformat{que}{\savehyperref{#1}{Question~\ref*{#1}}}
\newrefformat{op}{\savehyperref{#1}{Open Problem~\ref*{#1}}}
\newrefformat{fn}{\savehyperref{#1}{Footnote~\ref*{#1}}}
\allowdisplaybreaks
\usepackage{graphicx}
\usepackage{subcaption}
\usepackage{tabularx}
\usepackage{amsmath}
\usepackage[normalem]{ulem}

\newcommand{\algshort}{\textnormal{MPG-DPO}\xspace}

\usepackage{geometry}
\usepackage{lastpage}
\jmlrheading{1}{2024}{1-\pageref    {LastPage}}{4/00}{10/00}{Anonymous Authors}{}

\ShortHeadings{multidimensional preference-guided online direct preference optimization}{Zhang, Huang, Niu, Ishida, Sugiyama}
\firstpageno{1}

\begin{document}

\title{Sample-efficient Personalization of Large Language Model Alignment\\ with Multidimensional User Preference Data}

\author{\name Zhen-Yu Zhang \email zhen-yu.zhang@riken.jp \\
       \addr Center for Advanced Intelligence Project, RIKEN, Japan
       \AND
       \name Jerry Huang \email jerry.huang@mila.quebec \\
       \addr Center for Advanced Intelligence Project, RIKEN, Japan\\
       Université de Montréal, Canada\\
       Mila - Quebec AI Institute, Canada
       \AND
       \name Gang Niu \email gang.niu@riken.jp \\
       \addr Center for Advanced Intelligence Project, RIKEN
       \AND
       \name Takashi Ishida \email ishi@k.u-tokyo.ac.jp \\
       \addr Center for Advanced Intelligence Project, RIKEN, Japan\\
       Graduate School of Frontier Sciences, The University of Tokyo, Japan
       \AND
       \name Masashi Sugiyama \email sugi@k.u-tokyo.ac.jp \\
       \addr Center for Advanced Intelligence Project, RIKEN, Japan\\
       Graduate School of Frontier Sciences, The University of Tokyo, Japan
       }

\editor{}

\maketitle

\begin{abstract}
Preference alignment has become a central tool to align \emph{large language models}~{(LLMs)} with human values and desired behavior, and its online variant can further advance their capabilities beyond what static data alone offer by repeatedly generating and labeling new responses with human feedback.
But \emph{costly}, \emph{time-consuming} annotations limit its scalability, making sample efficiency essential, and existing methods typically seek to improve it by expanding coverage of the response space or selecting pairs with the highest uncertainty for comparison.
Although they are effective, these methods quantify uncertainty from the perspective of the LLM, which may differ from the uncertainty perceived by the user: a pair can appear highly uncertain to the LLM even when past comparisons from the user already imply a clear ordering.
In this paper, we propose \emph{multidimensional preference-guided online direct preference optimization}~{(\algshort{})}, which leverages low-cost multidimensional evaluations from automated verifiers on the user historical comparisons to construct a personalized preference cone guiding response selection for online preference alignment.
Specifically, it first discards responses that are confidently dominated under the estimated user preference, then selects two responses that are competitive under the retained preference cone yet exhibit the strongest unresolved disagreement. 
Theoretical analysis shows that, with high probability, the fraction of queries spent on cone-dominated responses vanishes as learning progresses.
Experiments across multiple LLM families show that \algshort{} outperforms strong contenders under matched query budgets.
\end{abstract}


\section{Introduction}
\label{sec:introduction}

Preference alignment has achieved remarkable success in aligning \emph{large language models}~{(LLMs)} with human preferences and desired behaviors~\citep{christiano2017deep,ouyang2022training}. 
To continually push the boundaries of model capabilities, preference alignment is typically extended to an \emph{online} process, where models repeatedly generate new responses for human evaluation~\citep{guo2024direct,calandriello2024human,xie2024exploratory}. 
However, the high cost and long turnaround of the annotations required pose a major barrier to scaling this continuous feedback loop. Therefore, improving sample efficiency in online preference alignment is of critical importance.

\begin{figure*}[t]
  \centering
  \includegraphics[width=0.93\textwidth,trim={0.50cm 4.30cm 0.28cm 6.20cm},clip]{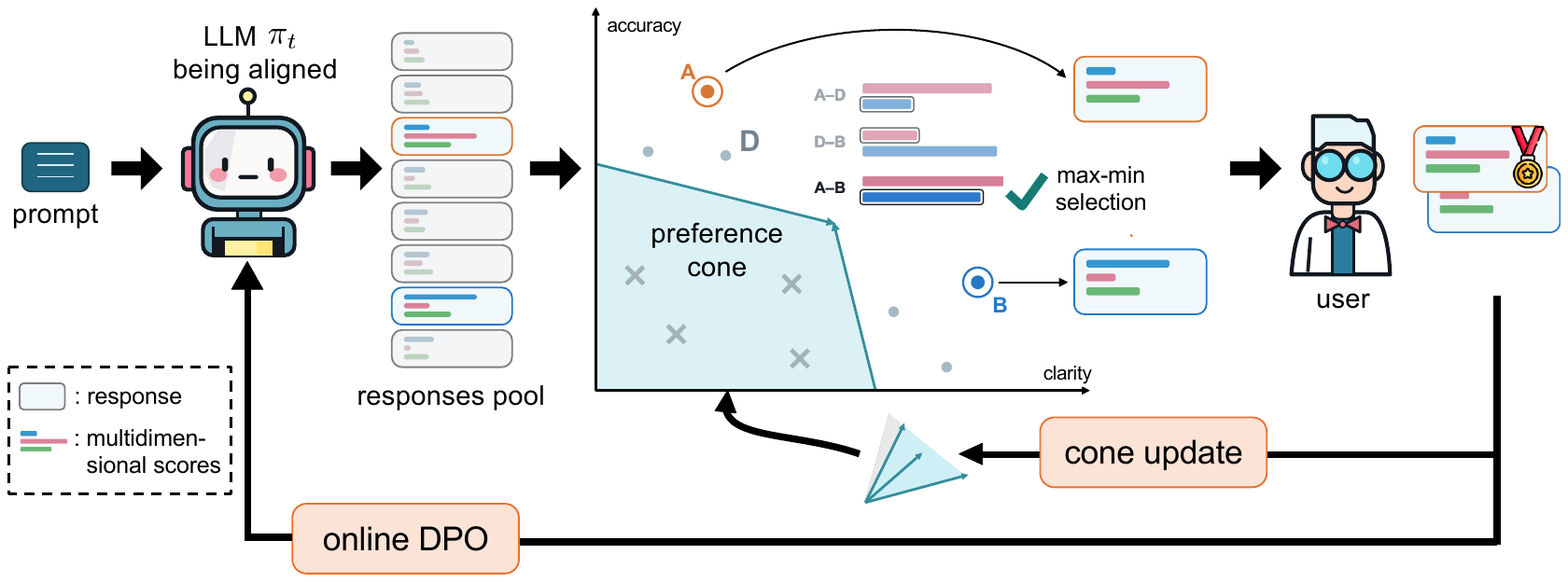}
  \caption{\textbf{Summary of the proposed \algshort{} approach.} Given an input prompt, candidate responses are sampled and filtered using the preference cone, after which a pair is selected to maximize the weaker of two opposing preference margins. The selected responses are evaluated and used to update the cone and policy.}
  \vspace{-2mm}
  \label{fig:illustration}
\end{figure*}

Existing approaches can be broadly divided into two categories. 
The first focuses on online exploration to improve policy coverage, increasing the sampling probability of responses that are poorly covered by previous data~\citep{xie2024exploratory,cen2024value,chen2025avoiding}. 
The second line of work relies on active preference alignment, which first generates a pool of candidate pairs and then selects the most informative pair for comparison in each interaction~\citep{das2024active,lin2025activedpo,belakaria2025sharpe}. 
These active methods typically rank candidates using information criteria measured from the model perspective, such as reward uncertainty or gradient-based metrics.

While effective at optimizing response generation or pair selection, these methods mainly rely on \emph{model-perspective} information, often querying feedback even when past comparisons made by the user already settle the ordering within a pair.
For example, in code generation, previous choices may show that the user is willing to trade some clarity for a sufficiently large gain in accuracy.
A new pair may still exhibit high reward uncertainty even when one response offers too little accuracy gain to justify its loss of clarity, making the comparison carry little \emph{user-perspective} information despite appearing uncertain to the model.
In a related direction, constitutional AI reduces repeated human supervision via high-level guiding principles~\citep{bai2022constitutional,huang2024collective}, but these principles are \emph{hand-written} and shared across users. 
To capture personalized trade-offs, specific dimensions like accuracy and clarity can fortunately be evaluated automatically, allowing them to naturally serve as low-cost multidimensional scores. 
This raises a question: \emph{How can low-cost multidimensional scores be leveraged to capture personalized preferences and improve sample efficiency?}

Using such multidimensional information is non-trivial because the scores are \emph{heterogeneous} and can hardly be directly aggregated. The insightful AMPLe~\citep{oh2025comparison} uses multi-objective rewards through Bayesian scalarization, but its uncertainty-driven queries may compare responses already dominated by other alternatives. We instead treat each historical comparison as revealing a \emph{preferred direction} of the user in the sub-objective space, and these directions form a personalized preference \emph{Pareto cone}~\citep{censor1977pareto,rame2023rewarded,ma2026selective}. The key challenge is to estimate this cone from historical data and use it to identify response pairs whose preference remains unresolved.

In this paper, we use low-cost multidimensional evaluations to improve sample efficiency in online preference alignment.
Each response is scored on attributes such as correctness and generation cost by low-cost or task-specific verifiers that are substantially cheaper than human evaluators. 
We construct a personalized preference cone from historical user comparisons and leverage its geometry to distinguish comparisons that are already resolved from those that represent genuinely unresolved trade-offs. 
We then select two competitive responses that \emph{exhibit the strongest unresolved disagreement among the retained preference directions}. 
This preserves the exploration of potentially valuable responses while directing costly feedback toward comparisons that provide new information.

We propose \emph{multidimensional preference-guided online direct preference optimization}~{(\algshort{})}, which uses the user preference cone to guide response selection at each round of online DPO. 
Specifically, at each round, the current policy generates a pool of on-policy responses, and a ridge-regression estimator predicts their multidimensional score vectors together with confidence bounds. 
We eliminate responses that are confidently dominated by another response in the sampled pool and select a pair for which different preference directions favor different responses. 
After obtaining the preference label, we update the preference cone and the policy. 
Theoretical analysis shows that, with high probability, only a slowly growing number of selections are cone-dominated, so wasted queries vanish as learning progresses.
Experiments show the effectiveness of the proposed method.

\section{Preliminaries}
\label{sec:problem-formulation}

In this section, we introduce the preliminaries and notations of our proposed algorithm.

\subsection{Online Preference Alignment}

In this part, we introduce the notations for online preference alignment. Let $\Pi$ be a class of LLM policies, where each $\pi\in\Pi$ maps a prompt $\x\in\X$ to a distribution $\pi(\cdot\mid\x)$ over the response space $\Y$. At each round $t$, a prompt $\x_t\sim\rho$ is drawn from a distribution $\rho$ over $\X$. Given $\x_t$, we sample two responses $y_t\sim\pi_t^{(1)}(\cdot\mid\x_t)$ and $y_t'\sim\pi_t^{(2)}(\cdot\mid\x_t)$, where $\pi_t^{(1)},\pi_t^{(2)}\in\Pi$ are the sampling policies used at round $t$. A preference oracle then returns a label $o_t\in\{-1,+1\}$, where $o_t=+1$ indicates that $y_t$ is preferred to $y_t'$, denoted by $y_t\succ y_t'$, and $o_t=-1$ indicates the reverse preference. 

For any policy $\pi\in\Pi$, we define its KL-regularized value as
\begin{equation*}
\label{eq:regularized-policy-value}
J_\beta(\pi) = \mathbb E_{\x\sim\rho,\,y\sim\pi(\cdot\mid\x)} \left[ r^*(\x,y) - \beta\log \frac{\pi(y\mid\x)}{\pi_{\rm ref}(y\mid\x)} \right],
\end{equation*}
where $r^*$ is the reward function, $\pi_{\rm ref}$ is a fixed reference policy and $\beta>0$ is the KL-regularization coefficient. Let $\pi_\beta^*\in\argmax_{\pi\in\Pi}J_\beta(\pi)$ denote an optimal KL-regularized policy.

The objective $J_\beta$ depends on the unknown reward function, whereas the available feedback consists of preference labels. \emph{Direct preference optimization}~(DPO) instead learns directly from preference pairs without fitting a separate reward model~\citep{rafailov2023direct}. We consider the online DPO setting~\citep{calandriello2024human,guo2024direct}, in which preference pairs are collected during the iterative training process. Given the comparison $(\x_t,y_t,y_t')$ and its observed preference label $o_t$ at round $t$, we perform the stochastic-gradient update
\begin{equation}
\pi_{t+1}=\pi_t-\eta_t\left.\nabla_{\pi}\left[-\log\sigma\left(\beta \ o_t\left[\log\frac{\pi(y_t\mid\x_t)}{\pi_{\rm ref}(y_t\mid\x_t)}-\log\frac{\pi(y_t'\mid\x_t)}{\pi_{\rm ref}(y_t'\mid\x_t)}\right]\right)\right]\right|_{\pi=\pi_t},
\label{eq:dpo}
\end{equation}
where $\eta_t>0$ is the learning rate.

\subsection{Preference Cone}

In this part, we introduce the notations of user preference cone. 

\begin{myDef}[Preference Matrix and Preference Cone]
\label{def:PD}
Let $K$ denote the number of preference directions and $d$ the number of sub-objectives. Given a preference matrix $\mathbf{W} \in \mathbb{R}^{K \times d}$, the associated preference cone is defined as
\[ C = \{ \mathbf{x} \in \mathbb{R}^d \mid \mathbf{W} \mathbf{x} \geq \bm{0} \}. \]
\end{myDef}

\begin{myDef}[Pareto $C$-Optimality]
\label{def:C-dom}
For any $\x, \y \in \mathbb{R}^d$, we say that $\x$ strictly $C$-dominates $\y$, denoted by $\x \succ_C \y$, if $\x - \y \in C$ and $\mathbf W(\x-\y)\neq\bm0$. An action is said to be \emph{Pareto \(C\)-optimal} if no other action strictly \(C\)-dominates it. We define the set of Pareto \(C\)-optimal actions as
\begin{equation*}
    P^*_{C} = \{ i \in \{1,\ldots,N\} \mid \nexists\, j \in \{1,\ldots,N\}\setminus\{i\}:\ j \succ_C i \}.
\end{equation*}
Here, $N$ is the number of actions. For notational convenience, for any two actions \(i,j\in\{1,\ldots,N\}\), we write \(i\succ_C j\) as shorthand for \(\x_i\succ_C\x_j\).
\end{myDef}

We assume that, at every round, each non-optimal response has an optimal alternative with a uniformly positive advantage across all preference directions available before the query, while allowing ties in other comparisons. This condition excludes non-optimal responses with only boundary dominators.
\begin{myAssum}[Uniform Positive Preference Gap]
\label{ass:uniform-round-gap}
Let $\mathbf w_{t-1,k}^\top$ denote the $k$-th row of $\mathbf W_{t-1}$ available before query $t$, and let $\x_{t,i}$ denote the latent score vector of action $i$ in the current candidate pool. All optimal sets in this assumption are evaluated on this pool using $C_{t-1}$. For each $i\notin P_{C_{t-1}}^{*}$, let $\mathcal D_t(i)=\{j\in P_{C_{t-1}}^{*}:j\succ_{C_{t-1}}i\}$ be nonempty and define its \emph{preference gap} as
\[
\Delta_{t,i}=\max_{j\in\mathcal D_t(i)}\min_{k\in\{1,\ldots,K_{t-1}\}}\frac{\mathbf w_{t-1,k}^\top(\x_{t,j}-\x_{t,i})}{U_{t-1,k}-L_{t-1,k}},
\]
where $L_{t-1,k}=\sum_{q=1}^{d}\min(0,\mathbf w_{t-1,k}[q])$, $U_{t-1,k}=\sum_{q=1}^{d}\max(0,\mathbf w_{t-1,k}[q])$, and $[q]$ indexes coordinates. We assume $\Delta_{t,i}\geq\Delta>0$ uniformly over all rounds $t$ and actions $i\notin P_{C_{t-1}}^{*}$.
\end{myAssum}


\section{Our Approach}
\label{sec:adaptive-cone-routing}

In this section, we present \emph{multidimensional preference-guided online direct preference optimization}~{(\algshort{})}. We first formulate the learning problem and then present our algorithm, which constructs a user preference cone from historical data to guide response pair selection for sample efficiency. Finally, we analyze the theoretical property of the proposed algorithm.

\subsection{Problem Formulation}
\label{subsec:online-problem-formulation}

In this part, we formulate the learning problem.

\textbf{Multidimensional score.}
We assume that each prompt--response pair $(\x,y)$ has an underlying score vector $\mu(\x,y)=\theta^{*\top}\phi(\x,y)\in[0,1]^d$. The observed multidimensional score is
\begin{equation*}
\z_t(\x,y)=\theta^{*\top}\phi(\x,y)+\epsilon_t(\x,y),\qquad \norm{\phi(\x,y)}_2\leq1,\qquad \norm{\theta^*}_F\leq S,
\end{equation*}
where $\phi(\x,y)\in\R^p$ is the prompt--response feature vector, $\theta^*\in\R^{p\times d}$ is unknown and fixed across rounds, and $\epsilon_t(\x,y)\in\R^d$ is a conditionally zero-mean, $\sigma_z$-sub-Gaussian noise vector, where $\sigma_z \ge 0$. We omit the prompt argument in $\phi$, $\mu$, $\z_t$, and $\epsilon_t$ when it is clear from the context.

\textbf{Learning protocol and performance measure.} In the online preference alignment setting, the learner observes multidimensional scores for the queried responses and receives pairwise preference feedback at each round. At each round $t$, the learner proceeds as follows:
\begin{list}{\arabic{enumi}.}{\usecounter{enumi}\setlength{\leftmargin}{1.5em}\setlength{\labelsep}{0.5em}\setlength{\itemsep}{0pt}\setlength{\parsep}{0pt}\setlength{\topsep}{0pt}\setlength{\partopsep}{0pt}}
\item Sample $N$ candidate responses from $\pi_t(\cdot\mid\x_t)$ and select a pair of indices $B_t=\{i_t,j_t\}$.
\item Observe the multidimensional score vectors of the selected responses and their preference label.
\item Update the estimated preference cone and the LLM policy.
\end{list}

The selection procedure is evaluated by whether a designated response from the selected pair belongs to the optimal set induced by the preference cone. Specifically, for any index $a_t\in B_t$, we define the cumulative indicator regret as
\vspace{-4mm}
\begin{align*}
\operatorname{Regret}^{01}(T)
&=\sum_{t=1}^T\mathbb I\{a_t\notin P_{C_{t-1}}^*\}.\label{eq:dynamic-indicator-regret}
\end{align*}

\subsection{Preference Cone Construction}
\label{subsec:adaptive-cone}

In this part, we construct the preference cone from historical comparisons. Each comparison reveals a user preference through the multidimensional score difference from the less preferred response to the preferred one. This difference defines a \emph{preference direction}, and non-negative linear combinations of these directions form the personalized preference cone. Since the feedback may be noisy, we screen new directions against existing preferences and retain only those that pass the consistency check.

Each comparison $(y_t,y_t')$ with label $o_t$ yields a preference direction by orienting the multidimensional score difference toward the preference label. Since scores and preference labels may be noisy, we use the previous preference matrix to discard a direction only when every row supports the opposite preference by a margin exceeding a threshold $\tau\geq0$. The retained directions generate the cone
\begin{equation}
    \vspace{-1mm}
C_t=\operatorname{cone}\left\{o_s\frac{\mathbf d_s}{\|\mathbf d_s\|_2}:1\leq s\leq t,\ \min_{1\leq k\leq K_{s-1}}\mathbf w_{s-1,k}^{\top}\left(-o_s\frac{\mathbf d_s}{\|\mathbf d_s\|_2}\right)\leq\tau\right\},
\label{eq:generated-preference-cone}
\end{equation}
where $\mathbf d_s=\z_s(y_s)-\z_s(y_s')$ is the observed multidimensional score difference, and $\operatorname{cone}$ denotes the set of all nonnegative linear combinations of the retained preference directions.

Given $C_t$, we obtain the preference matrix $\mathbf W_t$ using the double-description method~\citep{motzkin1953double,fukuda1995double}\footnote{A well-established implementation of the method is available through the Python package \texttt{pycddlib}.}. Specifically, we compute unit-norm generators of the vectors having nonnegative inner products with every point in $C_t$ and use their transposes as the rows of $\mathbf W_t$:
\begin{equation}
\vspace{-1mm}
\left\{\sum_{k=1}^{K_t}\lambda_k\mathbf w_{t,k}:\lambda_k\geq0\right\}=\left\{\mathbf w\in\R^d:\mathbf w^\top\mathbf x\geq0,\ \forall\mathbf x\in C_t\right\}.
\label{eq:preference-matrix}
\end{equation}
The resulting matrix gives $C_t=\{\mathbf x\in\R^d:\mathbf W_t\mathbf x\geq\bm0\}$. For computational efficiency, our implementation limits the number of retained rules, while the theoretical analysis uses the full cone.

\subsection{Response Pair Selection}
\label{subsec:preference-guided-sampling}

In this part, we use the current preference cone to guide response pair selection. The core idea is to \emph{select a pair of responses such that neither is $\mathcal{C}$-dominated by any other candidate, while their comparison reveals a trade-off unresolved by the current preference rules}. Each new query thus contributes a candidate preference direction in the multidimensional score space.

\textbf{Estimating multidimensional score confidence bounds.} We estimate multidimensional scores by fitting a ridge regression model to previously observed noisy response scores:
\begin{equation*}
V_t=\lambda I_p+\sum_{s<t}\sum_{i\in B_s}\phi(y_s^i)\phi(y_s^i)^\top,\qquad \widehat\theta_t=V_t^{-1}\sum_{s<t}\sum_{i\in B_s}\phi(y_s^i)\z_s(y_s^i)^\top.
\end{equation*}
Here, $\lambda>0$ is the regularization parameter. 

For any two responses $y$, $y'$ to $\x_t$, the predicted normalized improvement along a unit-norm row $\mathbf w_{t-1,k}^\top$ of $\mathbf W_{t-1}$ and the associated confidence width are
\begin{equation*}
\widehat m_{t,k}(y\mid y')=\frac{\mathbf w_{t-1,k}^\top\widehat\theta_t^\top[\phi(y)-\phi(y')]}{U_{t-1,k}-L_{t-1,k}},\qquad c_{t,k}(y,y')=\frac{\alpha_t(\delta)\|\phi(y)-\phi(y')\|_{V_t^{-1}}}{U_{t-1,k}-L_{t-1,k}},
\end{equation*}
where $\alpha_t(\delta)=\sqrt{\lambda}S+\sigma_z\sqrt{d\left(2\log\frac{d}{\delta}+\log\frac{\det V_t}{\det(\lambda I_p)}\right)}$.

Let $m_{t,k}(y\mid y')$ denote the corresponding latent margin, obtained by replacing $\widehat\theta_t$ with $\theta^{*}$. With probability at least $1-\delta$, $|m_{t,k}(y\mid y')-\widehat m_{t,k}(y\mid y')|\leq c_{t,k}(y,y')$ simultaneously over rounds (See Lemma~\ref{lem:directional-confidence} in Appendix~\ref{app:theorem}). The resulting confidence bounds are
\begin{equation}
\operatorname{UCB}_{t,k}(y\mid y')=\widehat m_{t,k}(y\mid y')+c_{t,k}(y,y'),\qquad \operatorname{LCB}_{t,k}(y\mid y')=\widehat m_{t,k}(y\mid y')-c_{t,k}(y,y').
\label{eq:preference-guided-pair-confidence}
\end{equation}
On this event, a positive lower bound certifies an improvement from $y'$ to $y$ along direction $\mathbf w_{t-1,k}$. 

\textbf{Removing $C$-dominated responses.} In the following, we show how to select the two responses. Firstly, a response is removed if another response in the pool confidently $C_{t-1}$-dominates it, so that we can prevent the human from comparing a response that is already known to be inferior to another available response. The retained indices are
\begin{equation}
\mathcal N_t = \left\{i\in[N]:\nexists\,j\in[N]\setminus\{i\}\text{ such that }\min_{k\in[K_{t-1}]}\operatorname{LCB}_{t,k}(y_t^j\mid y_t^i)>0\right\}.
\label{eq:pool-admissibility}
\end{equation}
For a fixed pair $(y_t^j,y_t^i)$, Eq.~\eqref{eq:pool-admissibility} evaluates the least favorable rule for $y_t^j$ improving upon $y_t^i$. A positive value means that all rows of $\mathbf W_{t-1}$ support this ordering with positive lower confidence bounds, allowing $y_t^i$ to be discarded without a human comparison.

\begin{algorithm}[t]
\caption{Multidimensional Preference-Guided Online Direct Preference Optimization}
\label{alg:preference-guided-online-dpo}
\begin{algorithmic}[1]
\STATE Receive $\x_t$ and sample $y_t^i\sim\pi_t(\cdot\mid\x_t)$ for $i\in[N]$.
\STATE Fit ridge regression to past noisy scores and compute confidence bounds using $\mathbf W_{t-1}$ via~\eqref{eq:preference-guided-pair-confidence}.
\STATE Form $\mathcal N_t$ using~\eqref{eq:pool-admissibility} and select $(i_t,j_t)$ using~\eqref{eq:rule-tradeoff-score}.
\STATE Set $y_t=y_t^{i_t}$ and $y_t'=y_t^{j_t}$, observe $\z_t(y_t),\z_t(y_t')$ and query the preference label $o_t$.
\STATE Compute $\mathbf d_t=\z_t(y_t)-\z_t(y_t')$ and construct $C_t$ using~\eqref{eq:generated-preference-cone}.
\STATE Obtain $\mathbf W_t$ using~\eqref{eq:preference-matrix} and update the score predictor with the observed scores.
\STATE Update $\pi_{t+1}$ using $(\x_t,y_t,y_t',o_t)$ via~\eqref{eq:dpo}.
\end{algorithmic}
\end{algorithm}

\textbf{Response pair selection.} After filtering out confidently dominated responses under the preference cone, we select a pair from $\mathcal N_t$ that maximizes the weaker of the two opposing preference margins:
\begin{equation}
(i_t,j_t)\in\underset{i,j\in\mathcal N_t,\ i\neq j}{\argmax}\min\!\left\{\max_{k\in[K_{t-1}]}\operatorname{LCB}_{t,k}(y_t^i\mid y_t^j),-\min_{k\in[K_{t-1}]}\operatorname{UCB}_{t,k}(y_t^i\mid y_t^j)\right\}.
\label{eq:rule-tradeoff-score}
\end{equation}
The two terms are positive when some rule confidently favors $y_t^i$ and $y_t^j$, respectively.
A positive minimum indicates conflicting preferences across rules and measures the weaker side of the disagreement.
Maximizing it favors pairs with strong evidence on both sides of the trade-off. 
The algorithm is summarized in Alg.~\ref{alg:preference-guided-online-dpo}. In implementation, we resample when fewer than two candidates survive, skip cone updates for zero score differences, and retain the previous cone if its update yields a trivial dual.

\subsection{Theoretical Analysis}
\label{subsec:indicator-regret-theory}

In this part, we analyze the theoretical property of the proposed algorithm. 
We note that our theoretical analysis provides a supporting explanation of the filtering mechanism by bounding the frequency of cone-dominated selections under simplifying assumptions. 
The empirical evaluation further examines the benefits of adaptive cone construction and trade-off selection for downstream performance.

We introduce the following assumption that requires the queried pair to provide sufficient information about a comparison that establishes the non-optimality of each selected response.
\begin{myAssum}[Coverage of a dominating response]
\label{ass:dominator-coverage}
Let $q_t=\sum_{i\in B_t}\|\phi(y_t^i)\|_{V_t^{-1}}^2$ denote the aggregate feature uncertainty of the queried pair. There exists a constant $\kappa>0$, independent of $t$, such that for every selected $i\in B_t\setminus P_{C_{t-1}}^*$, some $j\in\mathcal D_t(i)$ attaining the maximum defining $\Delta_{t,i}$ satisfies
\begin{equation}
\|\phi(y_t^j)-\phi(y_t^i)\|_{V_t^{-1}}^2\leq\kappa q_t.
\label{eq:dominator-coverage}
\end{equation}
\end{myAssum}

We bound regret by controlling the score-estimation uncertainty of responses dominated under $C_{t-1}$.
\begin{myThm}
\label{thm:dynamic-cone-indicator-regret}
Under Assumptions~\ref{ass:uniform-round-gap} and~\ref{ass:dominator-coverage}, let $\lambda\geq1$ and $\delta\in(0,1)$. With probability at least $1-\delta$, \algshort{} satisfies, simultaneously for all generated query prefixes of length $T\geq1$ and any choices $a_t\in B_t$,
\begin{equation}
\operatorname{Regret}^{01}(T)\leq \frac{12\bar\alpha_T^2(\delta)\kappa p}{\Delta^2}\log\left(1+\frac{2T}{\lambda p}\right),
\label{eq:single-regret-bound}
\end{equation}
where $\bar\alpha_T(\delta)=\sqrt{\lambda}S+\sigma_z\sqrt{d\left(2\log\frac{d}{\delta}+p\log\left(1+\frac{2T}{\lambda p}\right)\right)}$.
\end{myThm}

\begin{myRemark}
Unlike the fixed-cone setting of~\citet{ma2026selective}, our preference cone is estimated online from noisy pairwise comparisons and updated after every query, and is used to actively filter dominated responses and select the next query pair for online DPO. Selection and evaluation use the same cone $C_{t-1}$, so a dominated response can survive filtering only if its score comparison remains uncertain. The simultaneous confidence bound and coverage assumption control these selections without a separate term for cone updates.
\end{myRemark}

Theorem~\ref{thm:dynamic-cone-indicator-regret} shows that, if queries continue indefinitely, the average regret vanishes as $T\to\infty$ with probability at least $1-\delta$. The full proof is in Appendix~\ref{app:theorem}, which also shows that the round-level regret, counting a round once whenever at least one selected response is non-optimal under $C_{t-1}$, satisfies the same bound; summing the indicator over both responses instead yields twice this bound.

\begin{myProp}[Avoidance of cone-dominated queries]
\label{prop:preference-query-efficiency}
Under the conditions of Theorem~\ref{thm:dynamic-cone-indicator-regret}, with fixed problem parameters and $\delta$, for every generated prefix of $T$ expensive queries, at least $T-O(\log^2 T)$ compare two Pareto-optimal responses in their respective candidate pools under the cones available before the queries, with probability at least $1-\delta$. If queries continue indefinitely, the fraction allocated to such non-dominated pairs approaches one.
\end{myProp}

\section{Experiments}
\label{sec:experiments}

In this section, we first introduce the experimental setup and then present the experimental results. Our experiments aim to answer the following three questions: \textbf{Q1}: does the proposed method enhance the policy across different LLM families; \textbf{Q2}: does the proposed preference-guided sampling make better use of the multidimensional information than other methods with additional access to it; and \textbf{Q3}: how sensitive is the proposed method to its key parameters.

\subsection{Experimental Setup}
\label{subsec:experimental-setups}

\textbf{Prompts.} We use the prompt-only split of \texttt{ultrafeedback-prompt} for online training, following the generation and training pipeline of the official XPO implementation in \texttt{trl}\footnote{\url{https://huggingface.co/docs/trl/xpo_trainer}}. All methods share this pipeline. For \algshort{}, we initialize the preference matrix using comparisons collected from the preference oracle on \texttt{ultrafeedback\_binarized}.

\textbf{Models.} We evaluate three model families: \texttt{Qwen2.5-7B-Instruct}, \texttt{Mistral-7B-Instruct-v0.3}, and \texttt{OLMo-2-1124-7B-Instruct}. Each reference policy is initialized from the corresponding policy checkpoint and remains frozen throughout training. Unless otherwise specified, \algshort{} samples $N=8$ candidate responses per prompt from the current policy. All methods use the same prompt stream, maximum response length, decoding parameters, optimizer, and preference-query budget. We train with LoRA adapters~\citep{lora} using rank $r=32$ and scaling parameter $\alpha=64$, and report results over three random seeds. $\phi$ is meanwhile implemented as an independent feature extractor, which we fix to use an 86M parameter frozen \texttt{microsoft/deberta-v3-base}\footnote{\url{https://huggingface.co/microsoft/deberta-v3-base}} model on the prompt-response pair.

\textbf{Datasets and evaluation metrics.} We report four metrics: (i) average reward from a held-out reward model on the \textsc{IMDb} sentiment generation task~\citep{imdb}, (ii) summarization win rate against human-written summaries on \textsc{TL;DR}~\citep{tldr}, (iii) length-controlled win rate on \textsc{AlpacaEval2}~\citep{alpacaeval2}, and (iv) win rate on \textsc{Arena-Hard}~\citep{arena-hard}.

\textbf{Preference oracle.} We use \texttt{Kimi-K3}~\citep{kimi3} to simulate human preference feedback. Each query uses a fixed judging template containing the input prompt and two candidate responses, whose order is randomized independently for each query. The oracle identifies the preferred response, yielding a binary preference label. Each oracle call counts as one preference query. Preference labeling and sub-objective scoring are performed by separate models.

\textbf{Sub-objective oracle.} We use five fixed evaluation dimensions adapted from \texttt{FLASK}~\citep{flask}: correctness, completeness, clarity, conciseness, and reasoning quality. The corresponding scoring rubrics are generated by prompting the preference oracle. A frozen \texttt{Prometheus2} evaluator (\texttt{prometheus-7b-v2.0})~\citep{prometheus2} applies these rubrics to assign an integer score from 1 to 5 for each dimension. The scores are normalized to $[0,1]$ to obtain $\z_t(y)$. 

\textbf{Contenders.} We compare all methods under the same preference-query budget. We compared our approach against state-of-the-art contenders with official code implementations. Online DPO samples an on-policy response pair and applies the standard DPO update. XPO incorporates the exploratory objective of~\citet{xie2024exploratory}, while APL performs active preference selection following~\citet{oh2026random}. 
APL+MD modifies the APL selection algorithm by using a weighted average of the reward margin and the sum of squared ridge-prediction confidence widths between pairs of responses.

\subsection{Main Results}
\label{subsec:main-results}

Table~\ref{tab:main-results} compares \algshort{} with state-of-the-art baselines across three LLM families under the same preference-query budget. The consistent improvements across model families support the effectiveness of preference-guided pair selection.

\begin{table}[t]
    \centering
    \caption{Main results across three 7B policies under the same preference-query budget. Higher is better for every metric. Each entry reports mean $\pm$ standard deviation over three seeds.}
    \vspace{-0.5\baselineskip}
    \label{tab:main-results}
    \resizebox{0.98\linewidth}{!}{
    \begin{tabular}{llcccc}
        \toprule
        \multirow{2}{*}{LLM} & \multirow{2}{*}{Method} & \textsc{IMDb} & \textsc{TL;DR} & \textsc{AlpacaEval2} & \textsc{Arena-Hard} \\
        & & Average Reward & Win Rate (\%) & {LC Win Rate (\%)} & {Win Rate (\%)} \\
        \midrule
        \multirow{5}{*}{\texttt{Qwen2.5-7B}}
        & Online DPO    & {81.6}{\tiny $\pm$ 0.55} & {63.38}{\tiny $\pm$ 0.24} & {59.03}{\tiny $\pm$ 0.18} & {54.68}{\tiny $\pm$ 0.17} \\
        & XPO           & {81.8}{\tiny $\pm$ 0.66} & {63.82}{\tiny $\pm$ 0.27} & {59.14}{\tiny $\pm$ 0.20} & {54.74}{\tiny $\pm$ 0.24} \\
        & APL           & {82.1}{\tiny $\pm$ 0.61} & {64.02}{\tiny $\pm$ 0.26} & {59.13}{\tiny $\pm$ 0.22} & {54.94}{\tiny $\pm$ 0.21} \\
        & APL+MD        & {82.3}{\tiny $\pm$ 0.63} & {64.34}{\tiny $\pm$ 0.26} & {59.41}{\tiny $\pm$ 0.27} & {55.08}{\tiny $\pm$ 0.23} \\
        & \algshort{}   & \textbf{83.2}{\tiny $\pm$ 0.64} & \textbf{65.31}{\tiny $\pm$ 0.29} & \textbf{60.71}{\tiny $\pm$ 0.22} & \textbf{56.04}{\tiny $\pm$ 0.23} \\
        \midrule
        \multirow{5}{*}{\texttt{Mistral-7B}}
        & Online DPO    & {77.4}{\tiny $\pm$ 0.54} & {61.76}{\tiny $\pm$ 0.22} & {56.76}{\tiny $\pm$ 0.18} & {53.44}{\tiny $\pm$ 0.18} \\
        & XPO           & {77.8}{\tiny $\pm$ 0.57} & {61.87}{\tiny $\pm$ 0.24} & {56.81}{\tiny $\pm$ 0.20} & {53.52}{\tiny $\pm$ 0.24} \\
        & APL           & {77.9}{\tiny $\pm$ 0.53} & {61.89}{\tiny $\pm$ 0.25} & {56.88}{\tiny $\pm$ 0.17} & {53.47}{\tiny $\pm$ 0.20} \\
        & APL+MD        & {78.6}{\tiny $\pm$ 0.57} & {62.74}{\tiny $\pm$ 0.24} & {57.19}{\tiny $\pm$ 0.20} & {53.78}{\tiny $\pm$ 0.24} \\
        & \algshort{}   & \textbf{79.9}{\tiny $\pm$ 0.59} & \textbf{63.63}{\tiny $\pm$ 0.26} & \textbf{58.28}{\tiny $\pm$ 0.22} & \textbf{54.52}{\tiny $\pm$ 0.27} \\
        \midrule
        \multirow{5}{*}{\texttt{OLMo-2-7B}}
        & Online DPO    & {77.2}{\tiny $\pm$ 0.49} & {61.64}{\tiny $\pm$ 0.22} & {56.53}{\tiny $\pm$ 0.19} & {53.14}{\tiny $\pm$ 0.21} \\
        & XPO           & {78.1}{\tiny $\pm$ 0.54} & {62.67}{\tiny $\pm$ 0.28} & {56.49}{\tiny $\pm$ 0.21} & {53.28}{\tiny $\pm$ 0.22} \\
        & APL           & {77.2}{\tiny $\pm$ 0.52} & {61.58}{\tiny $\pm$ 0.25} & {56.44}{\tiny $\pm$ 0.18} & {53.20}{\tiny $\pm$ 0.23} \\
        & APL+MD        & {78.0}{\tiny $\pm$ 0.58} & {62.44}{\tiny $\pm$ 0.27} & {56.88}{\tiny $\pm$ 0.21} & {53.41}{\tiny $\pm$ 0.26} \\
        & \algshort{}   & \textbf{78.8}{\tiny $\pm$ 0.58} & \textbf{63.31}{\tiny $\pm$ 0.26} & \textbf{57.73}{\tiny $\pm$ 0.20} & \textbf{54.18}{\tiny $\pm$ 0.22} \\
        \bottomrule
    \end{tabular}}
\vspace{-4mm}
\end{table}

We also assess the contribution of each component in \algshort{} through the following ablations:
\begin{list}{\arabic{enumi}.}{\usecounter{enumi}\setlength{\leftmargin}{1.5em}\setlength{\labelsep}{0.5em}\setlength{\itemsep}{0pt}\setlength{\parsep}{0pt}\setlength{\topsep}{0pt}\setlength{\partopsep}{0pt}}
\item Without cone filtering: We omit candidate removal and apply the trade-off selection criterion to all pairs in the full candidate pool. This variant assesses the benefit of the filtering mechanism.
\item Without trade-off selection: We retain cone filtering but select a pair uniformly at random from $\mathcal N_t$. This variant assesses the benefit of trade-off selection beyond filtering alone.
\item Fixed Pareto cone: We replace the learned cone with the positive orthant by setting $\mathbf W_t=I_5$ throughout training, while retaining the filtering and trade-off selection procedures. The score predictor and policy continue to update. This variant compares learned preference trade-offs with a fixed coordinate-wise preference relation.
\end{list}
The results in Table~\ref{tab:component-ablation} show that the full method outperforms each ablated variant, supporting the contribution of each component of the proposed algorithm. With Table~\ref{tab:main-results}, these findings answer \textbf{Q1}.

\begin{table}[t]
\centering
\caption{Component ablations on \texttt{Qwen2.5-7B-Instruct} under a fixed preference-query budget. Each entry reports mean $\pm$ standard deviation over three independent training seeds.}
    \vspace{-0.5\baselineskip}
\label{tab:component-ablation}
\resizebox{0.98\linewidth}{!}{
\begin{tabular}{lcccc}
\toprule
\multirow{2}{*}{Method Variant} & \textsc{IMDb} & \textsc{TL;DR} & \textsc{AlpacaEval2} & \textsc{Arena-Hard} \\
& Average Reward & Win Rate (\%) & {LC Win Rate (\%)} & {Win Rate (\%)} \\
\midrule
Without cone filtering      & {81.9}{\tiny $\pm$ 0.62} & {64.04}{\tiny $\pm$ 0.25} & {58.88}{\tiny $\pm$ 0.23} & {54.74}{\tiny $\pm$ 0.22} \\
Without trade-off selection & {82.1}{\tiny $\pm$ 0.54} & {64.12}{\tiny $\pm$ 0.24} & {59.02}{\tiny $\pm$ 0.19} & {54.82}{\tiny $\pm$ 0.26} \\
Fixed Pareto cone           & {82.2}{\tiny $\pm$ 0.59} & {64.43}{\tiny $\pm$ 0.26} & {59.23}{\tiny $\pm$ 0.31} & {55.02}{\tiny $\pm$ 0.28} \\
Full method                 & \textbf{83.2}{\tiny $\pm$ 0.64} & \textbf{65.31}{\tiny $\pm$ 0.29} & \textbf{60.71}{\tiny $\pm$ 0.22} & \textbf{56.04}{\tiny $\pm$ 0.23} \\
\bottomrule
\end{tabular}}
\vspace{-3mm}
\end{table}

\subsection{Comparison with Alternative Uses of Multidimensional Information}
\label{subsec:multidimensional-studies}

We compare \algshort{} with alternative pair-selection strategies that use the same multidimensional feedback. Feedback budgets and training protocol are fixed, with only the selection criterion varied.

We construct three information-enhanced contenders:
\begin{list}{\arabic{enumi}.}{\usecounter{enumi}\setlength{\leftmargin}{1.5em}\setlength{\labelsep}{0.5em}\setlength{\itemsep}{0pt}\setlength{\parsep}{0pt}\setlength{\topsep}{0pt}\setlength{\partopsep}{0pt}}
\item Scalar-Margin selects the pair with the largest predicted scalar reward difference from the pool.
\item APL+MD-Score selects the pair with the largest sum of squared ridge-prediction confidence widths across its two responses, prioritizing uncertainty in the estimated multidimensional scores.
\item APL+MD-Scalar fits a Bradley--Terry model~\citep{bradley1952rank} to historical five-dimensional score differences and preference labels. Treating the predicted scores as fixed, it selects the pair with the largest approximate information gain about the scalarization weights.
\end{list}
These strategies provide comparisons with preference-cone-guided selection based on scalar reward differences, score-estimation uncertainty, and uncertainty in preference weights.

All methods share the same prompt stream, predictor architecture, warm-up data, scoring and preference-query budgets, and DPO hyperparameters. Only queried responses receive oracle scores.
We evaluate \texttt{Qwen2.5-7B-Instruct} and \texttt{Mistral-7B-Instruct-v0.3} over three independent training seeds. The results in Table~\ref{tab:multidimensional-ablation} support the effectiveness of preference-cone-guided selection relative to these alternative uses of multidimensional feedback, answering \textbf{Q2}.

\begin{table}[t]
    \centering
    \caption{Comparison of multidimensional feedback utilization across two policy models under matched scoring and preference-query budgets, with $N=8$. 
    }
    \vspace{-0.5\baselineskip}
    \label{tab:multidimensional-ablation}
    \resizebox{0.98\linewidth}{!}{
    \begin{tabular}{llcccc}
        \toprule
        \multirow{2}{*}{LLM}
        & \multirow{2}{*}{Method}
        & \textsc{IMDb}
        & \textsc{TL;DR}
        & \textsc{AlpacaEval2}
        & \textsc{Arena-Hard} \\
        & & Average Reward & Win Rate (\%)
        & LC Win Rate (\%) & Win Rate (\%) \\
        \midrule
        \multirow{4}{*}{\texttt{Qwen2.5-7B}}
        & Scalar-Margin             & {81.7}{\tiny $\pm$ 0.53} & {63.48}{\tiny $\pm$ 0.27} & {58.67}{\tiny $\pm$ 0.24} & {54.74}{\tiny $\pm$ 0.21} \\
        & APL+MD-Score              & {81.9}{\tiny $\pm$ 0.57} & {63.87}{\tiny $\pm$ 0.23} & {59.02}{\tiny $\pm$ 0.26} & {54.82}{\tiny $\pm$ 0.26} \\
        & APL+MD-Scalar      & {82.1}{\tiny $\pm$ 0.59} & {64.11}{\tiny $\pm$ 0.23} & {59.21}{\tiny $\pm$ 0.23} & {54.98}{\tiny $\pm$ 0.20} \\
        & \algshort{}               & \textbf{83.2}{\tiny $\pm$ 0.64} & \textbf{65.31}{\tiny $\pm$ 0.29} & \textbf{60.71}{\tiny $\pm$ 0.22} & \textbf{56.04}{\tiny $\pm$ 0.23} \\
        \midrule
        \multirow{4}{*}{\texttt{Mistral-7B}}
        & Scalar-Margin             & {77.6}{\tiny $\pm$ 0.57} & {61.83}{\tiny $\pm$ 0.22} & {56.79}{\tiny $\pm$ 0.17} & {53.14}{\tiny $\pm$ 0.22} \\
        & APL+MD-Score              & {77.7}{\tiny $\pm$ 0.54} & {61.97}{\tiny $\pm$ 0.27} & {57.01}{\tiny $\pm$ 0.23} & {53.31}{\tiny $\pm$ 0.22} \\
        & APL+MD-Scalar      & {78.3}{\tiny $\pm$ 0.54} & {62.26}{\tiny $\pm$ 0.27} & {57.11}{\tiny $\pm$ 0.20} & {53.51}{\tiny $\pm$ 0.25} \\
        & \algshort{}  & \textbf{79.9}{\tiny $\pm$ 0.59} & \textbf{63.63}{\tiny $\pm$ 0.26} & \textbf{58.28}{\tiny $\pm$ 0.22} & \textbf{54.52}{\tiny $\pm$ 0.27} \\
        \bottomrule
    \end{tabular}}
\vspace{-4mm}
\end{table}

\subsection{Ablation Studies}
\label{sec:ablation}

We examine the sensitivity of \algshort{} to the evaluation models, the candidate-pool size, the number of retained preference rules, and the number of sub-objectives.

\textbf{Effectiveness of personalization.} We evaluate adaptation to different oracle-defined preferences using \texttt{Qwen2.5-7B} as the backbone. In each setting, one of \texttt{Kimi-K2.6}, \texttt{DeepSeek-V4.1-Flash}, or \texttt{Qwen3.8-27B} serves as the oracle, providing preference feedback during training and computing win rates on \textsc{AlpacaEval2} and \textsc{Arena-Hard} during evaluation.
As shown in Table~\ref{tab:eval-model-ablation}, \algshort{} achieves higher mean win rates than the evaluated baselines in all three settings, supporting its effectiveness in adapting to different oracle-defined preferences.

\begin{figure}[t]
    \centering
    \begin{subfigure}[t]{0.32\linewidth}
        \centering
        \includegraphics[width=\linewidth]{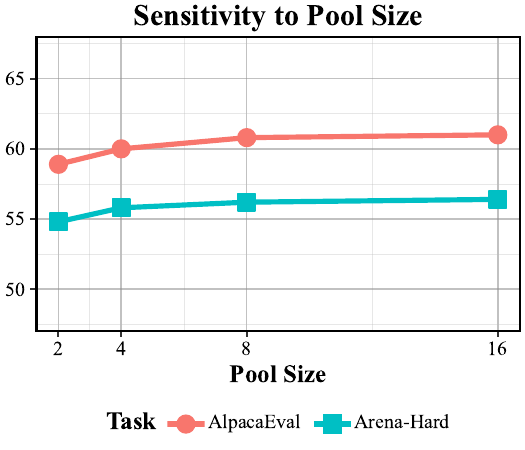}
        \caption{Pool size $N$.}
        \label{fig:pool_size}
    \end{subfigure}
    \hfill
    \begin{subfigure}[t]{0.32\linewidth}
        \centering
        \includegraphics[width=\linewidth]{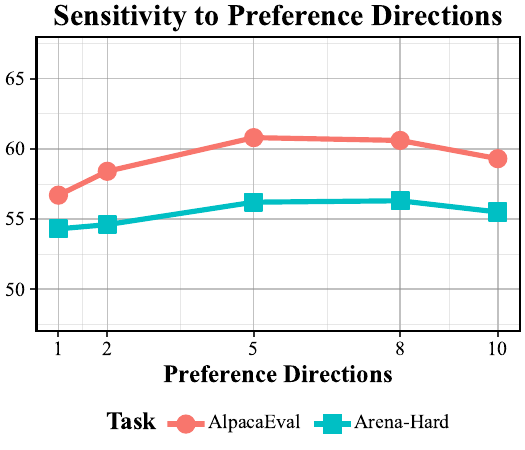}
        \caption{Number of preference rules $K$.}
        \label{fig:pref_rules}
    \end{subfigure}
    \hfill
    \begin{subfigure}[t]{0.32\linewidth}
        \centering
        \includegraphics[width=\linewidth]{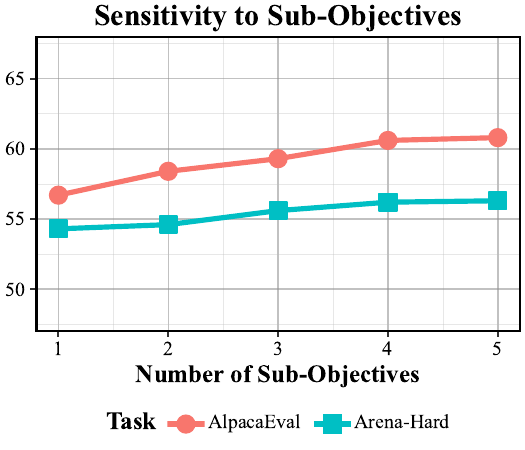}
        \caption{Number of sub-objectives $d$.}
        \label{fig:subobjectives}
    \end{subfigure}
        \vspace{-0.5\baselineskip}
    \caption{Sensitivity studies with reported win rates of tuned \texttt{Qwen2.5-7B} models versus the base checkpoint.}
    \label{fig:ablation-sensitivity}
\vspace{-4mm}
\end{figure}

\textbf{Candidate-pool size.} We vary $N\in\{2,4,8,16\}$ while holding the preference-query budget and DPO hyperparameters fixed. Larger pools provide more candidates for pair selection but increase response-generation cost. Figure~\ref{fig:pool_size} reports win rates on \textsc{AlpacaEval2} and \textsc{Arena-Hard}. Performance initially improves with pool size, with diminishing gains at larger values of $N$.

\textbf{Number of retained preference.} We vary the maximum number of rows retained in $\mathbf W_t$ to examine how the number of preferences affects pair selection. As shown in Figure~\ref{fig:pref_rules}, performance improves as the number of retained rules increases from small values, then declines slightly at larger settings. Intermediate settings achieve the highest win rates within the evaluated range.

\textbf{Number of sub-objectives.} We vary the number of sub-objectives by randomly selecting subsets of the \textsc{FLASK} evaluation dimensions. Figure~\ref{fig:subobjectives} shows that including more dimensions improves downstream performance within the evaluated range. Since the scoring rubrics are generated by the preference oracle, this trend may partly reflect the relevance of these dimensions to the target preferences. The interaction between the number of sub-objectives $d$ and the number of retained rules $K$ warrants further investigation.

\begin{table}[t]
    \centering
    \caption{Ablation study on the effectiveness of personalization.}
    \vspace{-0.5\baselineskip}
    \resizebox{\linewidth}{!}{
    \begin{tabular}{l|cc|cc||cc}
    \toprule
    \multirow{2}{*}{Method} & \multicolumn{2}{c|}{\texttt{Kimi-K2.6}} & \multicolumn{2}{c||}{\texttt{DeepSeek-V4.1-Flash}} & \multicolumn{2}{c}{\texttt{Qwen3.8-27B}} \\
    & \textsc{AlpacaEval2} & \textsc{Arena-Hard} & \textsc{AlpacaEval2} & \textsc{Arena-Hard} & \textsc{AlpacaEval2} & \textsc{Arena-Hard} \\
    \midrule
    Online DPO & {59.02}{\tiny $\pm$ 0.23} & {54.34}{\tiny $\pm$ 0.34} & {56.45}{\tiny $\pm$ 0.27} & {53.73}{\tiny $\pm$ 0.39} & {57.73}{\tiny $\pm$ 0.28} & {52.98}{\tiny $\pm$ 0.27} \\
    XPO        & {59.57}{\tiny $\pm$ 0.38} & {54.67}{\tiny $\pm$ 0.41} & {56.94}{\tiny $\pm$ 0.34} & {54.10}{\tiny $\pm$ 0.38} & {57.94}{\tiny $\pm$ 0.26} & {53.23}{\tiny $\pm$ 0.34} \\
    APL        & {59.08}{\tiny $\pm$ 0.28} & {54.45}{\tiny $\pm$ 0.25} & {56.75}{\tiny $\pm$ 0.23} & {53.85}{\tiny $\pm$ 0.33} & {57.93}{\tiny $\pm$ 0.32} & {53.44}{\tiny $\pm$ 0.31} \\
    \algshort{} & \textbf{60.45}{\tiny $\pm$ 0.33} & \textbf{55.98}{\tiny $\pm$ 0.36} & \textbf{57.04}{\tiny $\pm$ 0.40} & \textbf{54.31}{\tiny $\pm$ 0.38} & \textbf{59.26}{\tiny $\pm$ 0.29} & \textbf{54.33}{\tiny $\pm$ 0.33} \\
    \bottomrule
    \end{tabular}}
    \vspace{-2mm}
    \label{tab:eval-model-ablation}
\end{table}

These results answer \textbf{Q3}. Additional ablation results are provided in Appendix~\ref{app:add_results}.

\section{Related Work}
\label{sec:related-work}

This section reviews relevant prior work in detail.

\textbf{Exploration in online preference alignment.} Online preference alignment repeatedly generates responses from the current policy, obtains preference feedback, and updates the policy on the newly collected data~\citep{calandriello2024human}. A fruitful line of research aims to enhance policy-level coverage and generate high-value responses using a scalar implicit reward. XPO adds a principled exploration bonus to online DPO so that the policy can move beyond the support of the initial model and discover responses with higher latent reward~\citep{xie2024exploratory}. VPO similarly introduces optimism through value regularization, while recent theoretical work studies when preference-based exploration avoids poor dependence on the reward range~\citep{cen2024value,chen2025avoiding}. These methods primarily modify the DPO loss and can be seamlessly integrated into our algorithm. 

\textbf{Active preference selection.} A second line of work assumes a collection of candidate comparisons and actively decides which examples should receive costly preference labels~\citep{das2024active,lin2025activedpo,oh2026random}. APO formulates RLHF as a contextual preference bandit and allocates queries to statistically important prompt--response comparisons~\citep{das2024active}. ActiveDPO derives a selection score from uncertainty in the LLM's implicit reward difference and uses gradient projection and batch selection to make this criterion practical for nonlinear language models~\citep{lin2025activedpo}. Other work designs risk-adjusted active-learning criteria for preference optimization~\citep{belakaria2025sharpe}. These approaches rank candidates using scalar reward uncertainty while not exploiting the sub-objective trade-offs considered in our setting. Beyond preference learning, \citet{journals/aij/WangQZ2023} resolved a fundamental identifiability question through sound and complete orientation rules for local background knowledge in their latent-confounding setting. Their principled active causal discovery framework used accumulated knowledge to guide intervention selection, offering a complementary perspective on our preference-guided response-pair selection.

\textbf{Constitutional AI and rule-guided alignment.} Constitutional AI replaces repeated human supervision with a written set of high-level principles that guides self-critique, response revision, and AI-generated preference labels~\citep{bai2022constitutional}. \citet{lee2023rlaif} further study empirically whether an external language model can provide preference supervision at a quality comparable to human feedback, while collective constitutional AI obtains the constitution through a public deliberation process rather than solely from model developers~\citep{huang2024collective}. These methods demonstrate that explicit rules can make alignment supervision more scalable and interpretable. However, their constitutions are specified as natural-language principles and are typically shared across interactions. 

\textbf{Pareto learning.} The preference Pareto cone originates from~\citet{censor1977pareto} and has recently been used by \citet{rame2023rewarded} for reward interpolation and by \citet{ma2026selective} in multi-objective bandits. We adopt the same geometric abstraction but differ in both construction and use: our cone is estimated online from noisy pairwise comparisons via a consistency check, and is adaptively updated as new comparisons arrive to actively guide response selection.

\section{Conclusion}
\label{sec:conclusion}

In this paper, we proposed \algshort{}, an online preference alignment method that incorporates historical user preferences into response selection alongside model uncertainty for sample efficiency. Using low-cost multidimensional evaluations, the method constructs a personalized preference cone to filter confidently dominated responses and select pairs exhibiting unresolved trade-offs. This selection mechanism accounts for preference information already implied by past user comparisons. Under the stated assumptions, our analysis suggests that the number of selected cone-dominated responses grows slowly, so such queries make up a vanishing fraction as learning progresses. Experiments across multiple LLM families show improvements over strong baselines under matched preference-query budgets, supporting the effectiveness of preference-cone-guided selection.
\bibliography{./ref.bib}



\newpage

\appendix

\section{Additional Experimental Details}
\label{app:add_results}

\subsection{Setup Details}

\paragraph{Experimental Setup.} We run all online DPO variants for a fixed number of steps with a matched global batch size across all settings. Specific hyperparameters are listed in Table~\ref{tab:hyperparameters}.

\begin{table}[th]
    \centering
    \caption{Online DPO hyperparameters}
    \vspace{-0.5\baselineskip}
    \label{tab:hyperparameters}
    \small
    \begin{tabular}{l|c|l|c}
        \toprule
        \textbf{Hyperparameter} & \textbf{Value} & \textbf{Hyperparameter} & \textbf{Value} \\
        \midrule
        LoRA rank ($r$) & 32 & Epochs (query budget) & 3 \\
        LoRA ($\alpha$) & 64 & Global Batch Size & 64 \\
        Learning Rate & 5$\times$10\textsuperscript{-5} & Warm-up Ratio & 0.05 \\
        Optimizer & AdamW & Max Sequence Length & 256 \\
        Precision & \texttt{bf16} & Gradient Clipping & 1.0 \\
        \midrule
        \bottomrule
    \end{tabular}
\end{table}

 For \algshort{}, we use additional hyper-parameters of pool size $N=8$, $d=5$ sub-objectives and $K=8$. We further use $\tau=0.1$ and $\lambda=1.0$ by default. We use $\delta=0.1$, $S=1.0$, with $\alpha_t(\delta)$ computed as detailed in Section~\ref{subsec:preference-guided-sampling}.

 The human preference oracle is queried only once per update to determine the winning/losing response in the chosen pair. For $W_t$ initialization, a fixed amount of preference responses can be provided for initial computation, which can be conducted as an offline pre-processing step.

\paragraph{Computational Environment.} All experiments were conducted on a single machine with 4 NVIDIA H100 80GB GPUs. Experiments were run in an environment using CUDA version 12.8 and PyTorch 2.7.1.

\subsection{Additional Results}

\begin{wrapfigure}{R}{0.3\linewidth}
    \centering
    \includegraphics[width=\linewidth]{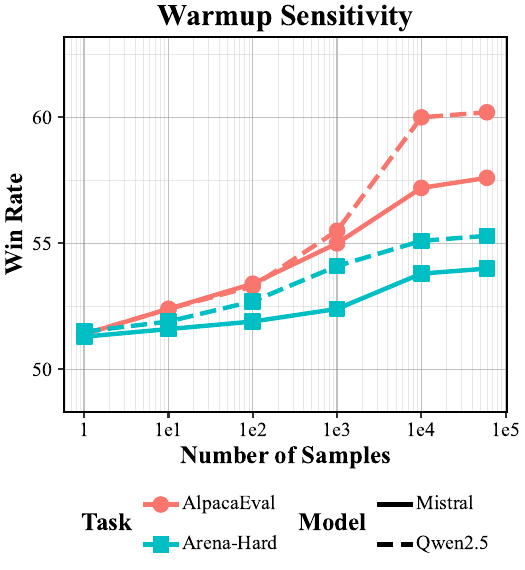}
    \vspace{-1.5\baselineskip}
    \caption{Number of queries for $W_t$ warm-up.}
    \label{fig:query_efficiency}
    \centering
    \includegraphics[width=\linewidth]{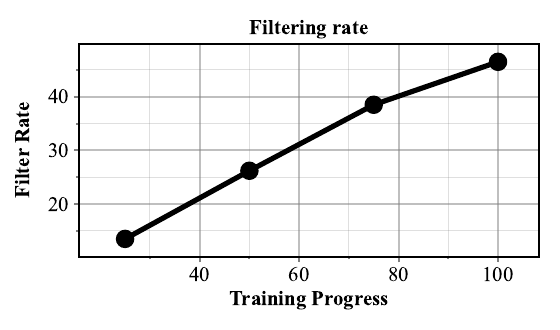}
    \vspace{-1.5\baselineskip}
    \caption{Filtering rate of responses during alignment.}
    \label{fig:auditing}
    \vspace{-1\baselineskip}
\end{wrapfigure}

\textbf{Query efficiency.} We next verify the importance of the warm-up step to construct $W_t$. Here, we use \texttt{Qwen2.5-7B-Instruct} and \texttt{Mistral-7B-Instruct-v0.3} and vary the amount of initial data necessary for $W_t$ construction. We re-use the same alignment procedure as described before. In \autoref{fig:query_efficiency}, we can see that a moderate amount of pre-labeled samples is important for properly initializing $W_t$. While the complete lack of any initial queries causes training to progress more slowly, which is sensible due to the need to construct $W_t$ completely from online pair-wise comparisons, using a moderate amount of warm-up labels can ensure high performance.

\textbf{Cone construction.} As a final verification, we also audit the cone construction process in Eq.~\ref{eq:generated-preference-cone}. We take the tuned model at different stages of tuning (25\%, 50\%, 75\%, and 100\%) and measure the filtering rate (removal of responses in the pool). We observe in \autoref{fig:auditing} that as the tuning progresses and both the cone as well as $W_t$ can encode increasing information about the model preferences, the model is able to filter out an increasing number of potentially less-informative responses within the pool.

\textbf{Varying Sub-Objective Oracle.} We also run a comparison by changing the use of the sub-objective oracle. As a simple replacement, we prompt a \texttt{Kimi-K2.6} using the same \textsc{Flask} objectives and templates used for Prometheus, obtaining scores from the new model for training and tuning with the setup described for the main experimental results.

\begin{table}[t]
    \centering
    \caption{Results using \texttt{Kimi-K2.6} sub-objective oracle. Each entry reports mean $\pm$ standard deviation over three seeds.}
    \vspace{-0.5\baselineskip}
    \label{tab:suboracle-kimi}
    \resizebox{0.98\linewidth}{!}{
    \begin{tabular}{llcccc}
        \toprule
        \multirow{2}{*}{LLM} & \multirow{2}{*}{Method} & \textsc{IMDb} & \textsc{TL;DR} & \textsc{AlpacaEval2} & \textsc{Arena-Hard} \\
        & & Average Reward & Win Rate (\%) & {LC Win Rate (\%)} & {Win Rate (\%)} \\
        \midrule
        \multirow{5}{*}{\texttt{Qwen2.5-7B}}
        & Online DPO    & {81.6}{\tiny $\pm$ 0.55} & {63.38}{\tiny $\pm$ 0.24} & {59.03}{\tiny $\pm$ 0.18} & {54.68}{\tiny $\pm$ 0.17} \\
        & XPO           & {81.8}{\tiny $\pm$ 0.66} & {63.82}{\tiny $\pm$ 0.27} & {59.14}{\tiny $\pm$ 0.20} & {54.74}{\tiny $\pm$ 0.24} \\
        & APL           & {82.1}{\tiny $\pm$ 0.61} & {64.02}{\tiny $\pm$ 0.26} & {59.13}{\tiny $\pm$ 0.22} & {54.94}{\tiny $\pm$ 0.21} \\
        & APL+MD        & {82.8}{\tiny $\pm$ 0.66} & {64.37}{\tiny $\pm$ 0.29} & {59.51}{\tiny $\pm$ 0.27} & {55.21}{\tiny $\pm$ 0.26} \\
        & \algshort{}   & \textbf{85.1}{\tiny $\pm$ 0.63} & \textbf{66.26}{\tiny $\pm$ 0.26} & \textbf{61.73}{\tiny $\pm$ 0.25} & \textbf{58.13}{\tiny $\pm$ 0.25} \\
        \bottomrule
    \end{tabular}}
\vspace{-4mm}
\end{table}

\section{Proof of Theorem~\ref{thm:dynamic-cone-indicator-regret}}
\label{app:theorem}

We first review the notations. At round $t$, the learner uses $C_{t-1}$ and $\mathbf W_{t-1}$ constructed from previous feedback, together with $V_t$ and $\widehat\theta_t$ fitted to noisy scores from rounds $s<t$. The current queried scores and preference label are observed after selection. The operational confidence bounds use
\[
\widehat m_{t,k}(y\mid y')=\frac{\mathbf w_{t-1,k}^\top\widehat\theta_t^\top[\phi(y)-\phi(y')]}{U_{t-1,k}-L_{t-1,k}},\qquad c_{t,k}(y,y')=\frac{\alpha_t(\delta)\|\phi(y)-\phi(y')\|_{V_t^{-1}}}{U_{t-1,k}-L_{t-1,k}},
\]
with $k\in[K_{t-1}]$ and $m_{t,k}$ obtained by replacing $\widehat\theta_t$ with $\theta^*$. All optimal sets are evaluated on the latent scores $\x_{t,i}=\mu(\x_t,y_t^i)$ in the current candidate pool. Regret, $\Delta_{t,i}$, and $\mathcal D_t(i)$ in Assumption~\ref{ass:uniform-round-gap} all use the same cone $C_{t-1}$ as selection. In particular, $P_{C_{t-1}}^*$ is evaluated on the current pool, before observing the current preference label. We use $q_t=\sum_{i\in B_t}\|\phi(y_t^i)\|_{V_t^{-1}}^2$ as defined in Section~\ref{subsec:indicator-regret-theory}.

\subsection{Technical Lemmas}
\label{app:lemmas}

In this part, we introduce necessary technical lemmas.

\begin{myLemma}[Simultaneous directional confidence bound]
\label{lem:directional-confidence}
Under the score model in Section~\ref{subsec:online-problem-formulation}, for any $\delta\in(0,1)$, with probability at least $1-\delta$, simultaneously for every round $t$, every $\mathbf v\in\R^p$, and every nonzero $\mathbf w\in\R^d$,
\begin{equation}
\frac{|\mathbf w^\top(\widehat\theta_t-\theta^*)^\top\mathbf v|}{\|\mathbf w\|_1}\leq\alpha_t(\delta)\frac{\|\mathbf w\|_2}{\|\mathbf w\|_1}\|\mathbf v\|_{V_t^{-1}}.
\label{eq:simultaneous-linear-confidence}
\end{equation}
Consequently, $\operatorname{LCB}_{t,k}(y\mid y')\leq m_{t,k}(y\mid y')\leq\operatorname{UCB}_{t,k}(y\mid y')$ simultaneously for all rounds, response pairs, and rows of $\mathbf W_{t-1}$.
\end{myLemma}

\textbf{Proof of Lemma~\ref{lem:directional-confidence}.}
Order the queried score vectors chronologically, using a fixed order within each pair. The corresponding feature vectors are known before each score is observed. By the self-normalized concentration inequality of~\citet[Theorem~1]{abbasi2011improved} and a union bound over the $d$ coordinates, with probability at least $1-\delta$, simultaneously for all $t$ and $q\in[d]$,
\[
\left\|\sum_{s<t}\sum_{i\in B_s}\phi(y_s^i)\epsilon_s(y_s^i)[q]\right\|_{V_t^{-1}}\leq\sigma_z\sqrt{2\log\frac{d}{\delta}+\log\frac{\det V_t}{\det(\lambda I_p)}}.
\]
Substitution of the noisy observations into the ridge estimator gives
\[
\widehat\theta_t-\theta^*=V_t^{-1}\left(\sum_{s<t}\sum_{i\in B_s}\phi(y_s^i)\epsilon_s(y_s^i)^\top-\lambda\theta^*\right).
\]
Combining the coordinate bounds in Frobenius norm and using $V_t\succeq\lambda I_p$ yields
\[
\|V_t^{1/2}(\widehat\theta_t-\theta^*)\|_F\leq\sigma_z\sqrt{d\left(2\log\frac{d}{\delta}+\log\frac{\det V_t}{\det(\lambda I_p)}\right)}+\lambda\|V_t^{-1/2}\theta^*\|_F\leq\alpha_t(\delta).
\]
Thus, for every $\mathbf v$ and $\mathbf w$, Cauchy--Schwarz implies
\[
|\mathbf w^\top(\widehat\theta_t-\theta^*)^\top\mathbf v|\leq\|\mathbf w\|_2\|V_t^{1/2}(\widehat\theta_t-\theta^*)\|_F\|\mathbf v\|_{V_t^{-1}}\leq\alpha_t(\delta)\|\mathbf w\|_2\|\mathbf v\|_{V_t^{-1}}.
\]
Dividing by $\|\mathbf w\|_1>0$ proves~\eqref{eq:simultaneous-linear-confidence}. Each unit-norm row satisfies
\[
U_{t-1,k}-L_{t-1,k}=\sum_{q=1}^d|\mathbf w_{t-1,k}[q]|=\|\mathbf w_{t-1,k}\|_1\geq\|\mathbf w_{t-1,k}\|_2=1.
\]
Taking $\mathbf w=\mathbf w_{t-1,k}$ and $\mathbf v=\phi(y)-\phi(y')$ gives $|m_{t,k}(y\mid y')-\widehat m_{t,k}(y\mid y')|\leq c_{t,k}(y,y')$. The confidence event is uniform over time and all directions, so adaptive candidate pools and preference matrices require no additional union bound.

\begin{myRemark}
Noisy scores and preference labels determine the realized cones through~\eqref{eq:generated-preference-cone}. Lemma~\ref{lem:directional-confidence} controls score-estimation error uniformly over time and directions, and the regret analysis uses the gap and coverage assumptions under the cones available before querying, without a separate concentration bound for preference labels. The resulting guarantee is relative to the learned cones; recovery of a latent preference cone or convergence of the true reward requires additional conditions.
\end{myRemark}

\begin{myLemma}[Filtering with the previous preference matrix]
\label{lem:lcb-elimination}
On the event of Lemma~\ref{lem:directional-confidence}, the filter evaluated using $\mathbf W_{t-1}$ retains every Pareto $C_{t-1}$-optimal response in the current pool. If $i\in\mathcal N_t$ and some response $j$ satisfies
\[
\min_{k\in[K_{t-1}]}\frac{\mathbf w_{t-1,k}^\top[\mu(y_t^j)-\mu(y_t^i)]}{U_{t-1,k}-L_{t-1,k}}\geq\Delta,
\]
then
\begin{equation}
\Delta\leq2\alpha_t(\delta)\|\phi(y_t^j)-\phi(y_t^i)\|_{V_t^{-1}}.
\label{eq:selected-gap-width}
\end{equation}
If this $j$ also satisfies~\eqref{eq:dominator-coverage}, then $\Delta^2\leq4\alpha_t^2(\delta)\kappa q_t$.
\end{myLemma}

\textbf{Proof of Lemma~\ref{lem:lcb-elimination}.}
If the filter removes $i$, another response $j$ has $\operatorname{LCB}_{t,k}(y_t^j\mid y_t^i)>0$ for every $k\in[K_{t-1}]$. Lemma~\ref{lem:directional-confidence} implies $\mathbf W_{t-1}[\mu(y_t^j)-\mu(y_t^i)]>\bm0$, and hence $j\succ_{C_{t-1}}i$. Thus no Pareto $C_{t-1}$-optimal response in the current pool is removed.

For a retained $i$ and a $j$ satisfying the stated margin condition, some $k\in[K_{t-1}]$ has $\operatorname{LCB}_{t,k}(y_t^j\mid y_t^i)\leq0$. The operational true margin $m_{t,k}$ uses $\mathbf w_{t-1,k}$, so Lemma~\ref{lem:directional-confidence} gives
\[
0\geq\operatorname{LCB}_{t,k}(y_t^j\mid y_t^i)\geq m_{t,k}(y_t^j\mid y_t^i)-2c_{t,k}(y_t^j,y_t^i)\geq\Delta-2c_{t,k}(y_t^j,y_t^i).
\]
Since $U_{t-1,k}-L_{t-1,k}=\|\mathbf w_{t-1,k}\|_1\geq1$,
\[
\Delta\leq\frac{2\alpha_t(\delta)\|\phi(y_t^j)-\phi(y_t^i)\|_{V_t^{-1}}}{U_{t-1,k}-L_{t-1,k}}\leq2\alpha_t(\delta)\|\phi(y_t^j)-\phi(y_t^i)\|_{V_t^{-1}}.
\]
Squaring and applying~\eqref{eq:dominator-coverage} proves the final claim. No current-round score is required to evaluate the filter; true margins are used only in the analysis.

\begin{myRemark}[A sufficient condition for coverage]
Assumption~\ref{ass:dominator-coverage} holds with $\kappa=4\gamma$ if, for a constant $\gamma>0$ independent of $t$,
\[
\max_{j\in[N]}\|\phi(y_t^j)\|_{V_t^{-1}}^2\leq\gamma q_t.
\]
For any selected response $i$ and any gap-achieving dominator $j$, the triangle inequality gives
\[
\|\phi(y_t^j)-\phi(y_t^i)\|_{V_t^{-1}}\leq\|\phi(y_t^j)\|_{V_t^{-1}}+\|\phi(y_t^i)\|_{V_t^{-1}}\leq2\sqrt{\gamma q_t}.
\]
The maximum ranges over the full candidate pool to cover every possible gap-achieving dominator under $C_{t-1}$. Coverage remains a separate assumption and is not implied by the trade-off selection criterion.
\end{myRemark}

\begin{myLemma}[Cumulative feature uncertainty]
\label{lem:pair-information-budget}
For $\lambda\geq1$, $|B_t|\leq2$, and $\|\phi(y_t^i)\|_2\leq1$,
\begin{equation}
\sum_{t=1}^Tq_t\leq3\log\frac{\det V_{T+1}}{\det V_1}\leq3p\log\left(1+\frac{2T}{\lambda p}\right).
\label{eq:cumulative-information-budget}
\end{equation}
\end{myLemma}

\textbf{Proof of Lemma~\ref{lem:pair-information-budget}.}
Let $H_t=V_t^{-1/2}(\sum_{i\in B_t}\phi(y_t^i)\phi(y_t^i)^\top)V_t^{-1/2}\succeq0$. Its eigenvalues $\mu_{t,1},\ldots,\mu_{t,p}$ are nonnegative, with $\operatorname{tr}(H_t)=\sum_{r=1}^p\mu_{t,r}=q_t$. Moreover, $V_t\succeq\lambda I_p$ gives
\[
0\leq q_t\leq\lambda^{-1}\sum_{i\in B_t}\|\phi(y_t^i)\|_2^2\leq2.
\]
Since $V_{t+1}=V_t^{1/2}(I_p+H_t)V_t^{1/2}$, we have
\[
\log\frac{\det V_{t+1}}{\det V_t}=\log\prod_{r=1}^p(1+\mu_{t,r})\geq\log(1+q_t)=\int_0^{q_t}\frac{du}{1+u}\geq\frac{q_t}{3}.
\]
The product inequality follows because all higher-order terms in its expansion are nonnegative; the last inequality uses $q_t\leq2$. Summing over $t$ telescopes the log determinants and proves the first bound. For the second, $V_1=\lambda I_p$ and
\[
\operatorname{tr}(V_{T+1})=\lambda p+\sum_{t=1}^T\sum_{i\in B_t}\|\phi(y_t^i)\|_2^2\leq\lambda p+2T.
\]
Applying the arithmetic--geometric mean inequality to the positive eigenvalues of $V_{T+1}$ yields
\[
\det V_{T+1}\leq\left(\frac{\operatorname{tr}(V_{T+1})}{p}\right)^p\leq\left(\lambda+\frac{2T}{p}\right)^p.
\]
Dividing by $\det V_1=\lambda^p$ and taking logarithms completes the proof.

\subsection{Proof of Theorem~\ref{thm:dynamic-cone-indicator-regret}}
\label{app:main_proof}

In this part, we provide a detailed proof of Theorem~\ref{thm:dynamic-cone-indicator-regret}. We work on the event of Lemma~\ref{lem:directional-confidence} and fix arbitrary choices $a_t\in B_t$. At every query with $a_t\notin P_{C_{t-1}}^*$, choose a gap-achieving $j\in\mathcal D_t(a_t)$ satisfying Assumption~\ref{ass:dominator-coverage}. Assumption~\ref{ass:uniform-round-gap} gives
\[
\min_{k\in[K_{t-1}]}\frac{\mathbf w_{t-1,k}^\top[\mu(y_t^j)-\mu(y_t^{a_t})]}{U_{t-1,k}-L_{t-1,k}}=\Delta_{t,a_t}\geq\Delta.
\]
Since $a_t$ survived the filter based on $\mathbf W_{t-1}$, Lemma~\ref{lem:lcb-elimination} and coverage yield $\Delta^2\leq4\alpha_t^2(\delta)\kappa q_t$. For any generated query prefix of length $T$, the determinant bound in Lemma~\ref{lem:pair-information-budget} gives $\alpha_t(\delta)\leq\bar\alpha_T(\delta)$ for $t\leq T$. Summing over nonoptimal selections and applying the same lemma yields
\[
\operatorname{Regret}^{01}(T)\leq\frac{4\bar\alpha_T^2(\delta)\kappa}{\Delta^2}\sum_{t=1}^Tq_t\leq\frac{12\bar\alpha_T^2(\delta)\kappa p}{\Delta^2}\log\left(1+\frac{2T}{\lambda p}\right).
\]
This proves~\eqref{eq:single-regret-bound} on the simultaneous confidence event, whose probability is at least $1-\delta$.

The choices $a_t$ were arbitrary on the same confidence event. In particular, choose a nonoptimal member whenever $B_t$ contains one. This gives the same bound for $\sum_{t=1}^T\mathbb I\{B_t\not\subseteq P_{C_{t-1}}^*\}$. Since each pair contains at most two responses, $\sum_{t=1}^T\sum_{i\in B_t}\mathbb I\{i\notin P_{C_{t-1}}^*\}$ is at most twice this count.

For fixed $\lambda,S,\sigma_z,\kappa,p,d,\Delta,\delta$, we have $\bar\alpha_T^2(\delta)=O(\log T)$. If queries continue indefinitely, the bound holds simultaneously for all $T$ on the same event, so $\operatorname{Regret}^{01}(T)/T\to0$ with probability at least $1-\delta$.

\subsection{Full Statement and Proof of Proposition~\ref{prop:preference-query-efficiency}}
\label{app:query-efficiency}

In this part, we provide the full statement of Proposition~\ref{prop:preference-query-efficiency} and corresponding proof. Under the conditions of Theorem~\ref{thm:dynamic-cone-indicator-regret}, let $\mathcal R_T(\delta)$ denote the right-hand side of~\eqref{eq:single-regret-bound}, and let $G_T=\sum_{t=1}^T\mathbb I\{B_t\subseteq P_{C_{t-1}}^*\}$ count queries whose two responses are Pareto-optimal in their respective pools under the cones available before querying. For each integer $M\geq1$, let $Q_M=\inf\{T\in\mathbb N:G_T\geq M\}$ be the number of expensive queries needed to collect $M$ such comparisons, with the infimum taken over generated query prefixes and $\inf\varnothing=+\infty$. With probability at least $1-\delta$, simultaneously for all generated query prefixes $T\geq1$, and for all integers $M\geq1$ such that at least $2M$ queries are generated,
\begin{equation}
\frac{G_T}{T}\geq1-\frac{\mathcal R_T(\delta)}{T},\qquad M\leq Q_M\leq M+\left\lceil\mathcal R_{2M}(\delta)\right\rceil\quad\text{if }\mathcal R_{2M}(\delta)\leq M.
\label{eq:preference-query-efficiency}
\end{equation}
If queries continue indefinitely, then for fixed problem parameters and $\delta$, $Q_M=M+O(\log^2 M)$ as $M\to\infty$.

\textbf{Proof of Proposition~\ref{prop:preference-query-efficiency}.}
We work on the same simultaneous confidence event as in Theorem~\ref{thm:dynamic-cone-indicator-regret}. For each round, choose $a_t\in B_t\setminus P_{C_{t-1}}^*$ when this set is nonempty, and choose any $a_t\in B_t$ otherwise. Since the theorem holds simultaneously for all choices of $a_t$, including these choices made for evaluation, we have
\[
T-G_T=\sum_{t=1}^T\mathbb I\{B_t\not\subseteq P_{C_{t-1}}^*\}=\sum_{t=1}^T\mathbb I\{a_t\notin P_{C_{t-1}}^*\}\leq\mathcal R_T(\delta).
\]
Dividing by $T$ proves the first inequality in~\eqref{eq:preference-query-efficiency}. Each round produces at most one counted comparison, so $Q_M\geq M$.

The bound $\mathcal R_T(\delta)$ is nondecreasing in $T$. If $\mathcal R_{2M}(\delta)\leq M$, then the integer $T=M+\lceil\mathcal R_{2M}(\delta)\rceil$ satisfies $T\leq2M$, and hence
\[
G_T\geq T-\mathcal R_T(\delta)\geq M+\lceil\mathcal R_{2M}(\delta)\rceil-\mathcal R_{2M}(\delta)\geq M.
\]
It follows that $Q_M\leq M+\lceil\mathcal R_{2M}(\delta)\rceil$. For fixed problem parameters and $\delta$, $\mathcal R_{2M}(\delta)=O(\log^2 M)=o(M)$, so, if queries continue indefinitely, the condition holds for all sufficiently large $M$, giving $Q_M=M+O(\log^2 M)$ on the same confidence event. No additional union bound is needed.

The count $G_T$ evaluates each comparison using its own round's candidate pool and the cone available before querying; previously collected comparisons are not reevaluated under later cones. The proposition quantifies query allocation to nondominated pairs, without assuming that all such comparisons have equal information value or establishing a label complexity for a target policy reward.



\end{document}